\documentclass[conference]{IEEEtran}
\usepackage{mathtools,bm,amsfonts,amssymb,algorithm,algorithmicx,caption,algpseudocode,mathrsfs,multirow,multicol,subcaption,booktabs,threeparttable,balance,tikz,breqn,dsfont,upgreek,lineno}

\usepackage[bookmarks=false]{hyperref}

\newcommand{\figwidth}{0.23}
\newcommand{\sysname}{\textsf{NAS} }
\newcommand{\sysnamenospace}{\textsf{NAS}}
\newcommand{\coname}{\textsf{Amadeus} }
\newcommand{\conamens}{\textsf{Amadeus}}

\newcommand{\ie}{\emph{i.e.}}
\newcommand{\eg}{\emph{e.g.}}

\newtheorem{definition}{Definition}
\DeclareMathOperator*{\minimize}{minimize}

\makeatletter
\newcommand*\bigcdot{\mathpalette\bigcdot@{.8}}
\newcommand*\bigcdot@[2]{\mathbin{\vcenter{\hbox{\scalebox{#2}{$\m@th#1\bullet$}}}}}
\makeatother

\usepackage{varwidth}
\DeclareCaptionFormat{centercaption}{%
  \begin{varwidth}{\linewidth}%
    \centering
    #1#2#3%
  \end{varwidth}%
}
  
\renewcommand{\vec}[1]{\bm{#1}}
\algnewcommand\algorithmicinput{\textbf{INPUT:}}
\algnewcommand\INPUT{\item[\algorithmicinput]}
\algnewcommand\algorithmicoutput{\textbf{OUTPUT:}}
\algnewcommand\OUTPUT{\item[\algorithmicoutput]}
\algnewcommand\algorithmicforeach{\textbf{for each}}
\algdef{S}[FOR]{ForEach}[1]{\algorithmicforeach\ #1\ \algorithmicdo}
 
\begin{document}
\bstctlcite{NAS:BSTcontrol}

\IEEEoverridecommandlockouts
\title{Attributed Sequence Embedding}
\author{
\IEEEauthorblockN{Zhongfang Zhuang, Xiangnan Kong, Elke, Rundensteiner}
\IEEEauthorblockA{
{Worcester Polytechnic Institute} \\
\{zzhuang, xkong, rundenst\}@wpi.edu}
\and
\IEEEauthorblockN{Jihane Zouaoui, Aditya Arora}
\IEEEauthorblockA{{Amadeus IT Group}\\
\{jihane.zouaoui, aditya.arora\}@amadeus.com}
}

\IEEEpubid{\begin{minipage}{\textwidth}\ \\[12pt] 
    978-1-7281-0858-2/19/\$31.00~\copyright~2019~IEEE
  \end{minipage}}

\IEEEpubidadjcol

\maketitle

\selectfont
\begin{abstract}
Mining tasks over sequential data, such as clickstreams and gene sequences, require a careful design of embeddings usable by learning algorithms. 
Recent research in feature learning has been extended to sequential data, where each instance consists of a sequence of heterogeneous items with a variable length. 
However, many real-world applications often involve \textit{attributed sequences}, where each instance is composed of both a sequence of categorical items and a set of attributes. 
In this paper, we study this new problem of \textit{attributed sequence embedding}, where the goal is to learn the representations of attributed sequences in an unsupervised fashion. 
This problem is core to many important data mining tasks ranging from user behavior analysis to the clustering of gene sequences. 
This problem is challenging due to the dependencies between sequences and their associated attributes. 
We propose a deep multimodal learning framework, called \sysnamenospace, to produce embeddings of attributed sequences. 
The embeddings are task independent and can be used on various mining tasks of attributed sequences. 
We demonstrate the effectiveness of our embeddings of attributed sequences in various unsupervised learning tasks on real-world datasets. 
\end{abstract}
\section{Introduction}
\label{intro}
Sequential data arise naturally in a wide range of applications~\cite{bechet2015sequence, miliaraki2013mind, wang2016unsupervised, wei2013effective}. Examples of sequential data include click streams of web users, purchase histories of online customers, and DNA sequences of genes. Different from conventional multidimensional data~\cite{pedersen1999multidimensional}, the sequential data~\cite{yang2003cluseq} are not represented as feature vectors of continuous values, but as sequences of categorical items with variable-lengths. 

Many real-world applications involve mining tasks over sequential data~\cite{wei2013effective, tajer2014outlying,wang2016unsupervised}. For example, in online ticketing systems, administrators are interested in finding fraudulent sequences from the clickstreams of users. In user profiling systems, researchers are interested in grouping purchase histories of customers into clusters. Motivated by these real-world applications, sequential data mining has received considerable attention in recent years~\cite{miliaraki2013mind, bechet2015sequence}. 

Sequential data usually requires a careful design of its embedding before being fed to data mining algorithms. One of the feature learning problems on sequential data is called sequence embedding~\cite{cho2014learning,sutskever2014sequence}, where the goal is to transform a sequence into a fixed-length embedding. 

\begin{figure}[t]
    \centering
    \includegraphics[width=0.9\linewidth]{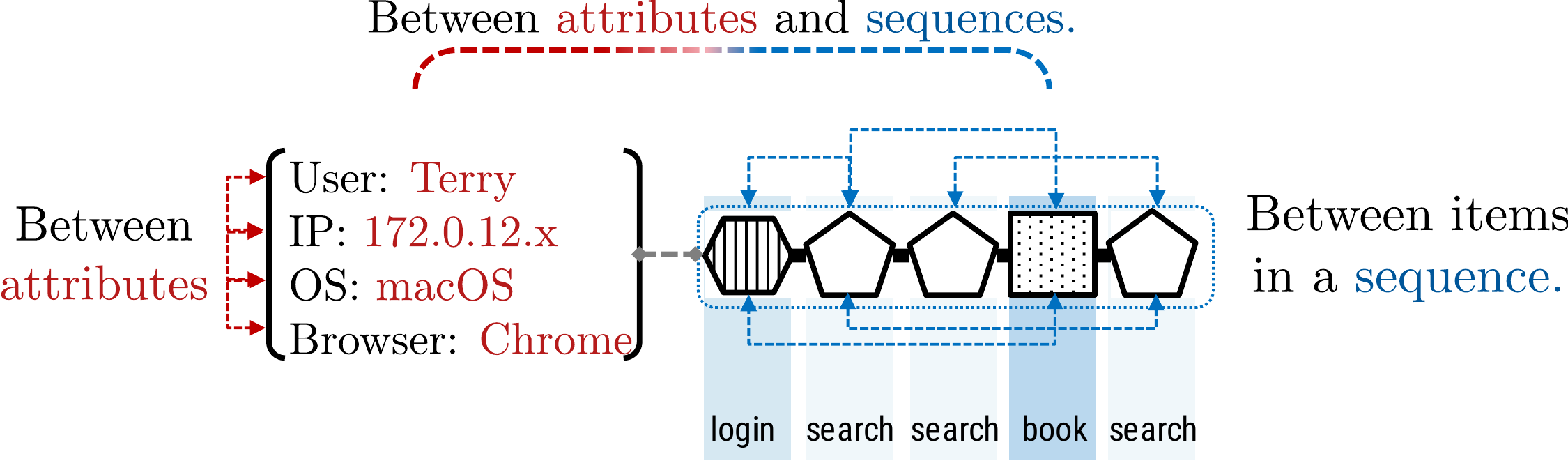}
    \vspace{-10pt}
    \caption{An example of attributed sequences. The three types of dependencies in an attributed sequence: {\color{blue} item} dependencies, {\color{red} attribute} dependencies and {\color{red}attribute}-{\color{blue}sequence} dependencies. }
    \label{fig-complexrel}
  \vspace{-15pt}
\end{figure}

Conventional methods on sequence embedding focus on learning from sequential data alone~\cite{kalchbrenner2013recurrent, cho2014learning, sutskever2014sequence, luong2015multi}. 
However, in many real-world applications, sequences are often associated with a set of attributes. We define such data as \textit{attributed sequences}, where each instance is represented by a set of \textit{attributes} associated with a \textit{sequence}. For example, in online ticketing systems as shown in Fig.~\ref{fig-complexrel}, each user transaction includes both a sequence of user actions (\textit{e.g.}, ``\texttt{login}'', ``\texttt{search}'' and ``\texttt{pick seats}'') and a set of attributes (\textit{e.g.}, ``\texttt{user name}'', ``\texttt{browser}'' and ``\texttt{IP address}'') indicating the context of the transaction. In gene function analysis, each gene can be represented by both a DNA sequence and a set of attributes indicating the expression levels of the gene in different types of cells. 
Motivated by the recent success in attributed graph embedding~\cite{gibert2012graph, perozzi2014deepwalk}, in this paper, we study the problem of attributed sequence embedding. 

\begin{figure}[t]
    \centering
    \begin{subfigure}{0.45\linewidth}
        \begin{subfigure}{1.1\linewidth}
          \centering
          \includegraphics[width=1\textwidth]{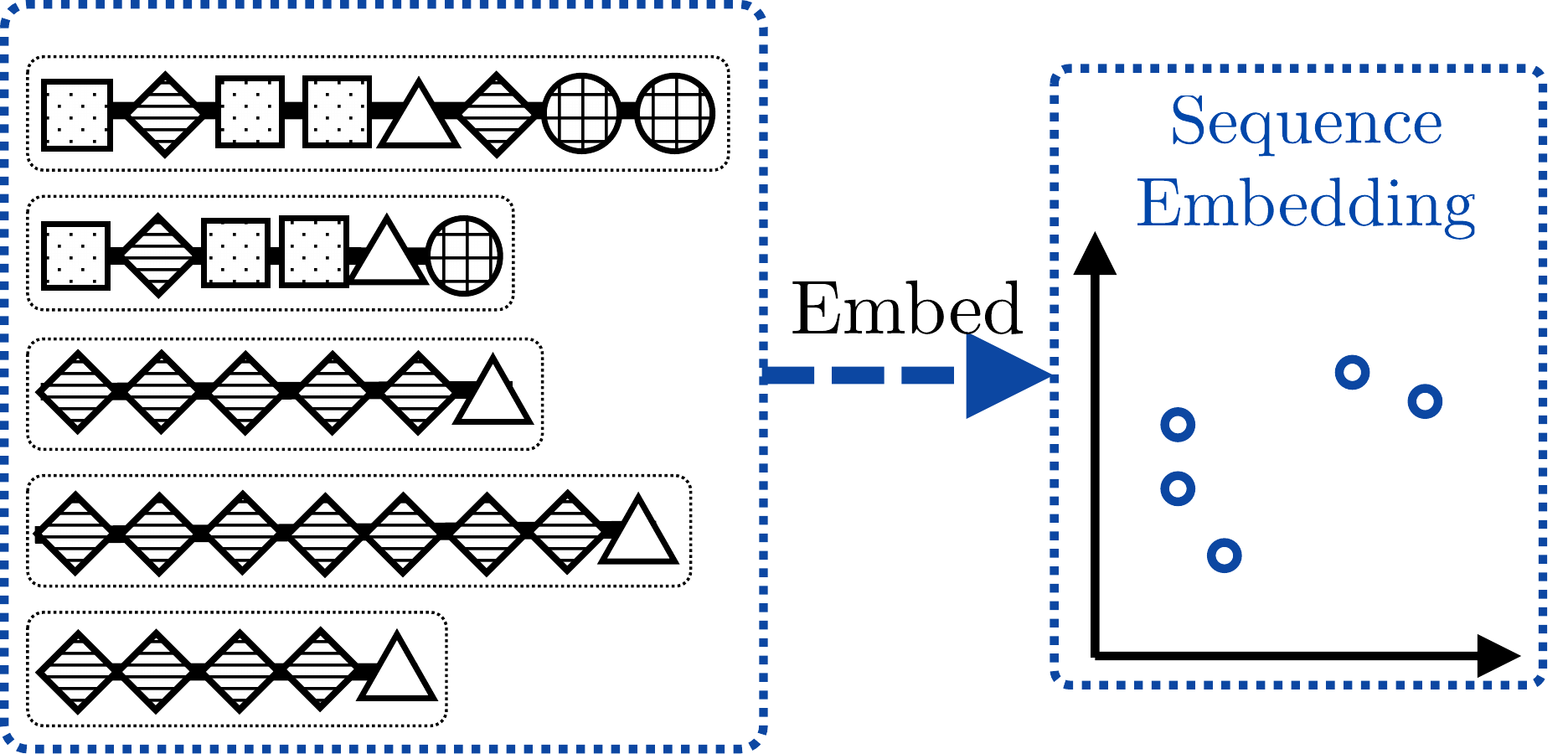}
          \caption{$\!\!$Sequence embedding~\cite{sutskever2014sequence}.}
          \label{fig-sota-seqonly}
         \vspace{5pt}
        \end{subfigure}
        \begin{subfigure}{1\linewidth}
          \centering
          \includegraphics[width=0.9\textwidth]{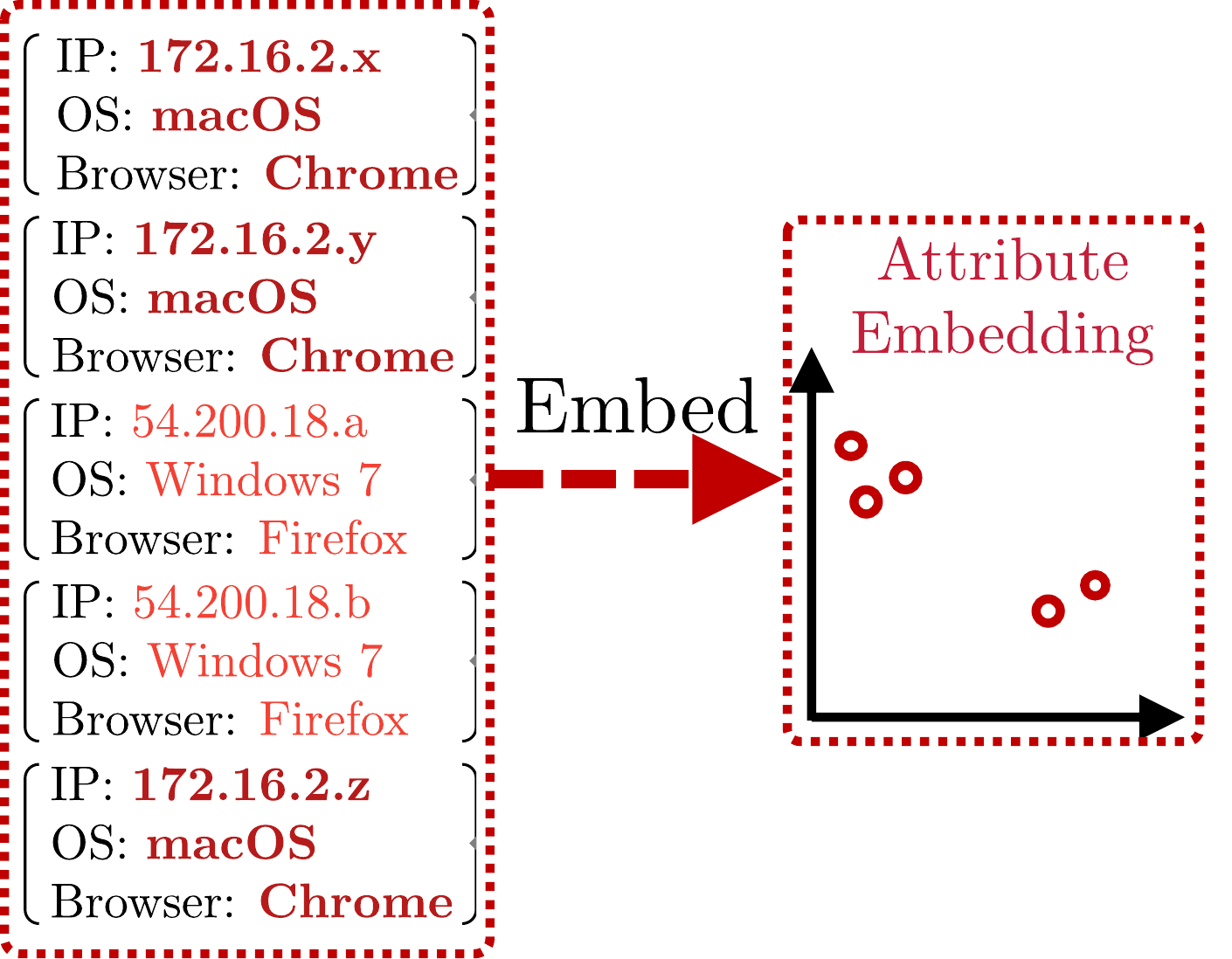}
          \caption{$\!\!\!$Attribute embedding\cite{wang2014generalized}.}
          \label{fig-sota-attr}
          \vspace{5pt}
        \end{subfigure}
        \begin{subfigure}{1\linewidth}
          \centering
          \includegraphics[width=0.95\textwidth]{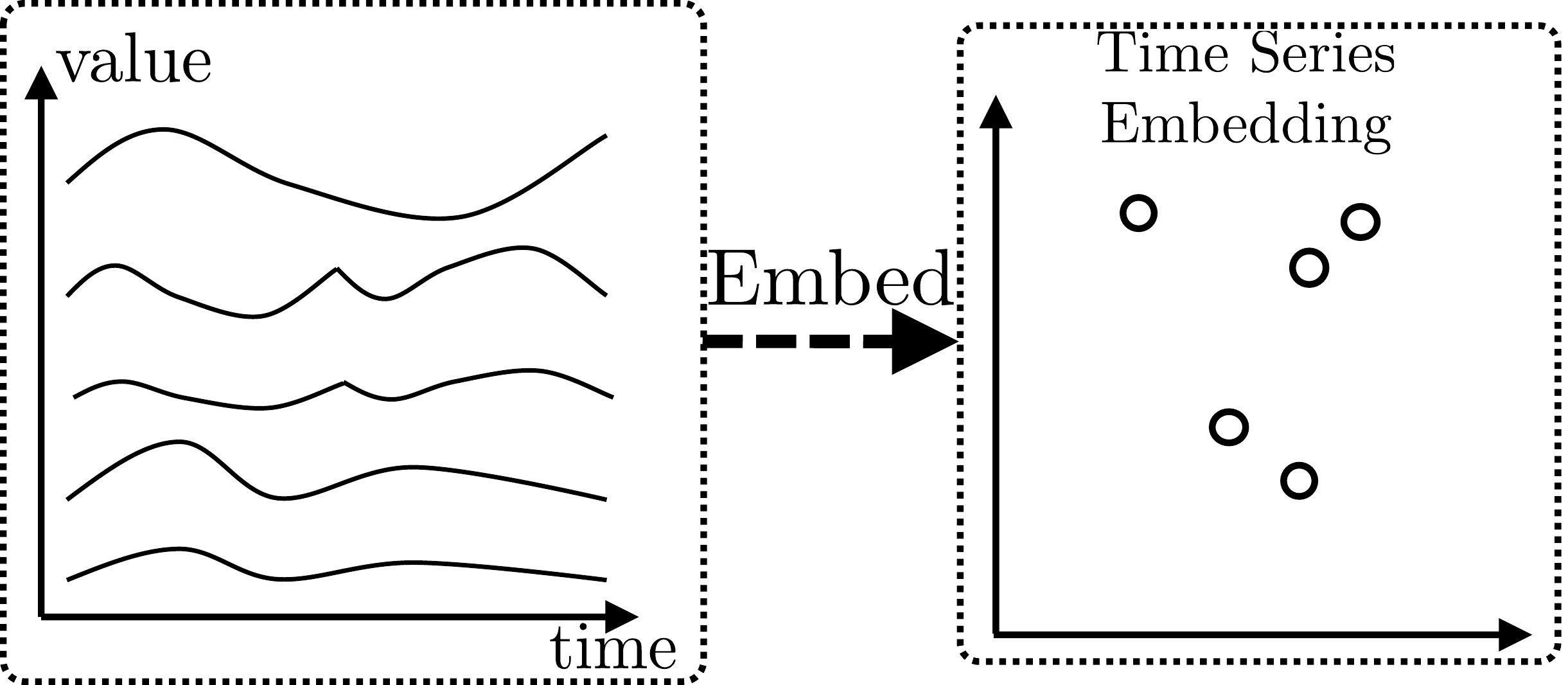}
          \caption{$\!\!$Time series embedding~\cite{khaleghi2016consistent}. }
          \label{fig-sota-ts}
        \end{subfigure}
    \end{subfigure}
    \hspace{5mm}
    \begin{subfigure}{0.40\linewidth}
    \centering
        \includegraphics[width=1\textwidth]{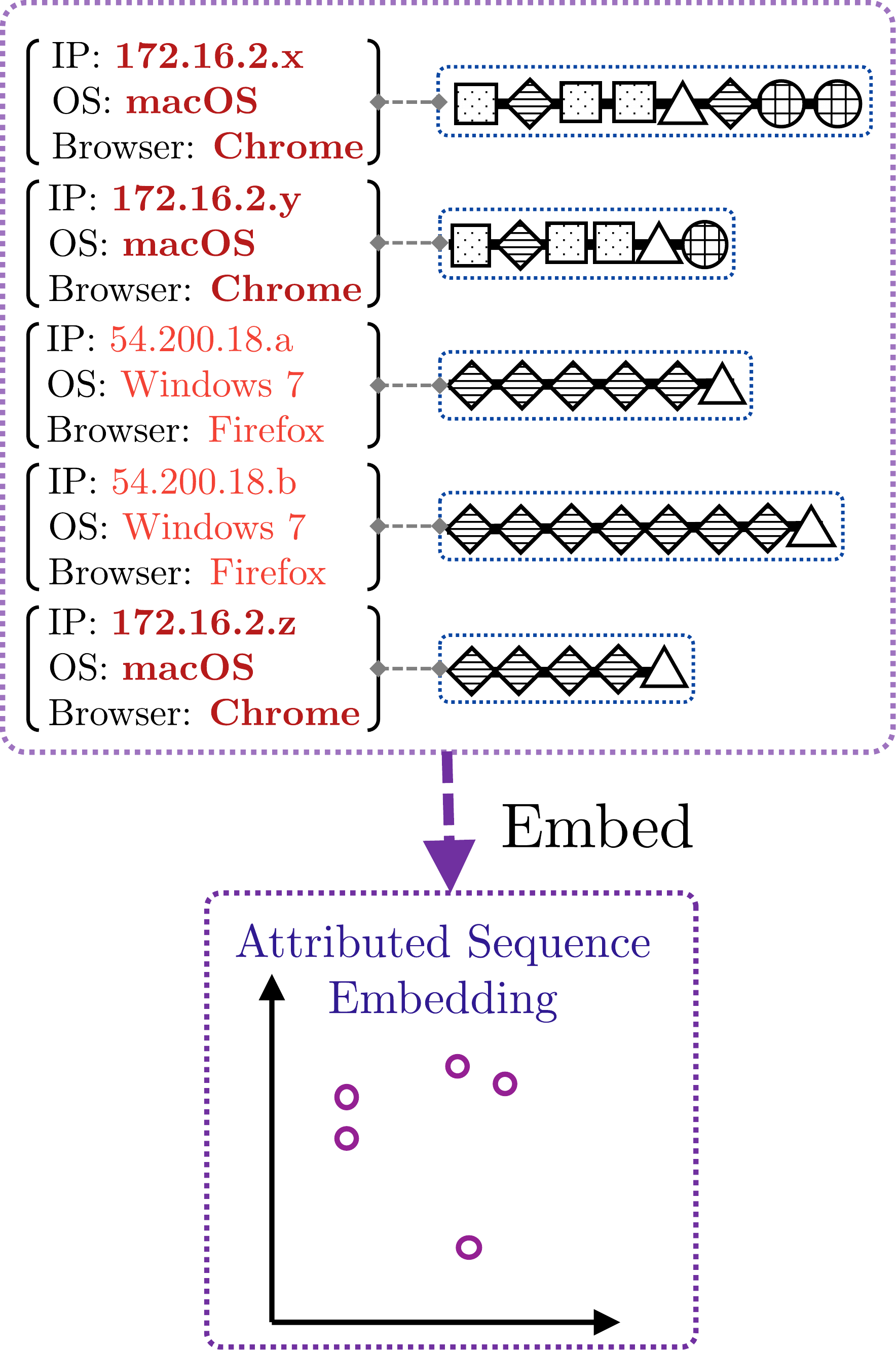}
        \caption{Attributed sequence embedding (this paper). }
        \label{fig-sota-nas}
    \end{subfigure}
    \caption{Comparison of different embedding problems. }
    \vspace{-10pt}
    \label{fig-as-setting}
\end{figure}

Building embedding for attributed sequences (as shown in Fig.~\ref{fig-sota-nas} corresponds to transforming an attributed sequence into a fixed-length embedding with continuous values. Different from the work in~\cite{zhuang2018one, zhuang2019amas}, we do not have labels for any attributed sequence instances in the embedding task. Sequence embedding problems are particularly challenging with additional attributes. In sequence embedding problems (as shown in Fig.~\ref{fig-sota-seqonly}, conventional methods focus on modeling the \emph{item dependencies}, {\ie}, the dependencies between different items within a sequence. However, in attributed sequences, the dependencies between items can be different if the sequence is observed under different contexts (attributes). Even the same ordering of the items can have different meanings if associated with different attribute values. 
In this paper, instead of building embeddings to model only the dependencies between items in each single sequence, we aim to model three types of dependencies in an attributed sequence jointly:  (1) \emph{item dependencies}, (2) \emph{attribute dependencies} ({\ie}, the dependencies between different attributes) and (3) \emph{attribute-sequence dependencies} ({\ie}, the dependencies between attributes and items in a sequence). 

Despite its relevance, the problem of producing attributed sequence embeddings in an unsupervised setting remains open. 
We summarize the major research challenges as follows:
\begin{enumerate}
\item \textbf{Heterogeneous Dependencies.}
The bipartite structure of attributed sequences poses unique challenges in feature learning. 
As shown in Fig.~\ref{fig-complexrel}, there exist three types of possible dependencies in an attributed sequence: item dependencies, attribute dependencies and attribute-sequence dependencies. 

\noindent \textit{\underline{Motivating Example 1.}} In Fig.~\ref{fig-as2vec}, we present an example of fraud detection from a user privilege management system in \conamens~\cite{amadeus}. This system logs each user session as an attributed sequence (denoted as $J_1 \sim J_5$). Each attributed sequence consists of a sequence of user's activities and a set of attributes derived from metadata values. The attributes ({\eg}, ``\texttt{IP}'', ``\texttt{OS}'' and ``\texttt{Browser}'') are recorded when a user logs into the system and remain unchanged during each user session. We use different shapes and colors to denote different user activities, {\eg}, ``\texttt{Reset password}'', ``\texttt{Delete a user}''. In real-world applications like this, the attributes and the associated sequences are already saved within one integrated record. An important step in this fraud detection system is to ``\textit{red flag}'' suspicious user sessions for potential security breaches. In Fig.~\ref{fig-as2vec}, we observe three groups of embeddings learned from the \coname application logs. For each group, we use a dendrogram to demonstrate the similarities between embeddings within that group. Neither of the embeddings using only sequences or only attributes detects any outliers due to the lacking of considerations of attribute-sequence dependencies. However, user session $J_5$ is discovered to be fraudulent using a learning algorithm that incorporates all three types of dependencies. 

\begin{figure}[t]
    \centering
    \includegraphics[width=0.7\linewidth]{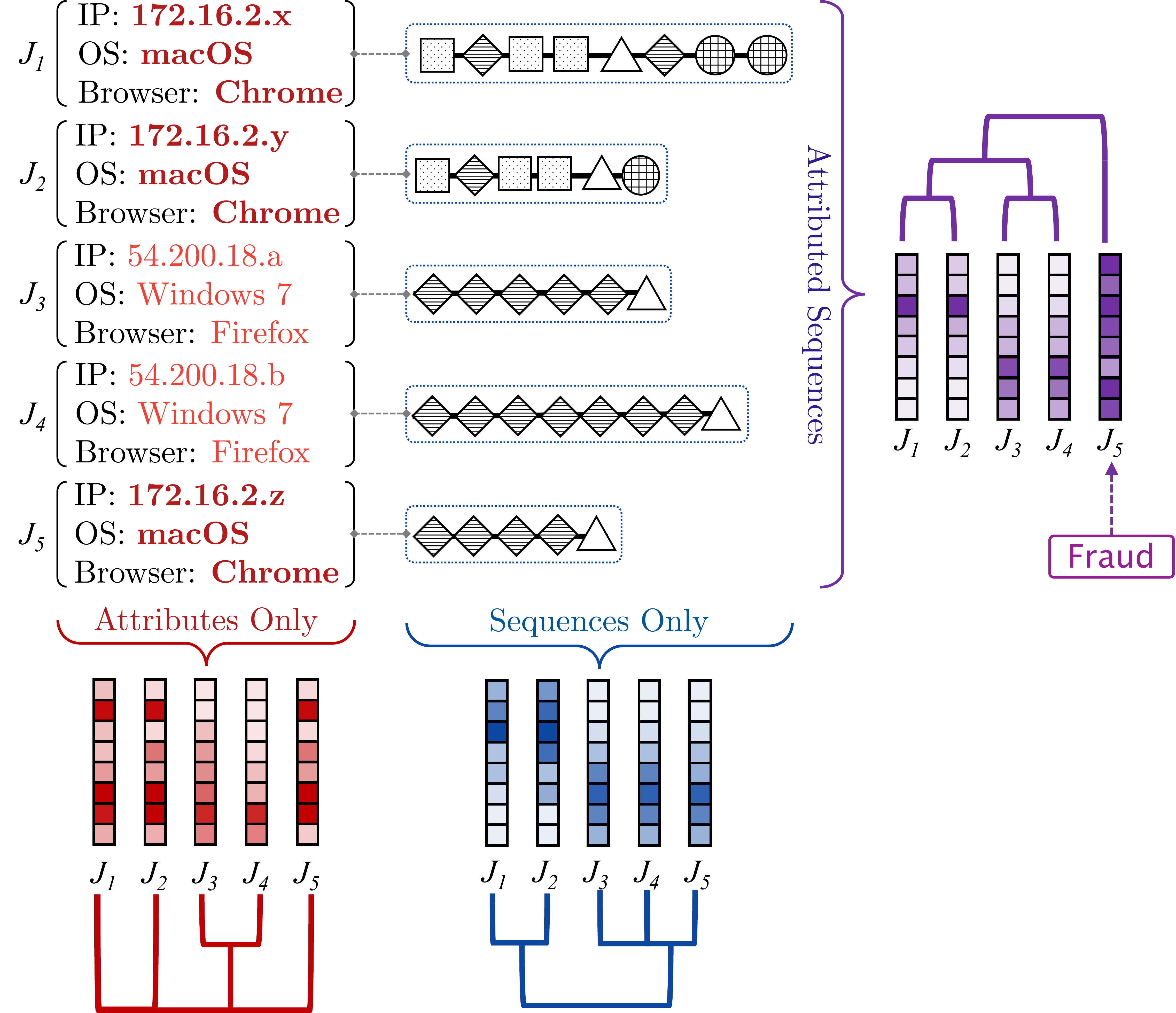}
    \caption{Dendrograms of embeddings learned from attributed sequences for fraud detection tasks. $J_5$ is a user committing fraud. However, it is considered a normal user session by the embedding generated using either only attributes or only sequences. $J_5$ can only be caught as a fraud instance using the embedding learned using both attributes and sequences.}
    \label{fig-as2vec}
    \vspace{-20pt}
\end{figure}
\item \textbf{Lack of Labeled Data. } With the continuously incoming volume of data and the high labor cost of manually labeling data, it is rare to find attributed sequences from many real-world applications with labels ({\eg}, \textit{fraud}, \textit{normal}) attached. Without proper labels, it is challenging to learn an embedding function that is capable of transforming attributed sequences into compact embeddings concerning the three types of dependencies. 

\noindent\textit{\underline{Motivating Example 2.}} Continuing with our Motivating Example 1, the \coname records user activities and their session metadata in the log files. Due to the large volume of entries and complex user sessions, the log files do not have labels depicting whether one user session is fraudulent or not. Only when an embedding function that is capable of transforming unlabeled user sessions $J_1 \sim J_5$ respecting the differences between them, an anomaly detection algorithm can identify $J_5$ as a fraudulent session.
\end{enumerate}

In this paper, we focus on the \textit{generic} problem of embedding attributed sequences in an unsupervised fashion. We propose a novel framework (called \sysnamenospace) using deep learning models to address the above challenges. This paper offers the following contributions: 
\begin{itemize}
\item {We study the problem of attributed sequence embedding without any labels available. }
\item We propose a framework and a training strategy to exploit the dependencies among the attributed sequences. %
\item We evaluate the embeddings generated by \sysname framework on real-world datasets using outlier detection tasks. We also conduct case studies of user behavior analysis and demonstrate the usefulness of \sysname in real-world applications. 
\end{itemize}

\section{Problem Formulation}
\label{prob-form}
\subsection{Preliminaries}
\begin{definition}[Sequence]{\rm
\label{plain-sequence}
Given a set of $r$ categorical items $\mathcal{I} = \{e_1,\!\cdots, e_r\}$, the $k$-th sequence in the dataset $S_k = \left( \alpha_k^{(1)}, \cdots, \alpha_k^{(l_k)} \right)$ is an ordered list of items, where $\alpha_k^{(t)} \in \mathcal{I}, \forall t = 1,\cdots, l_k$.}
\end{definition}
Different sequences can have a varying number of items. For example, the number of user click activities varies between different user sessions. 
The meanings of items are different in different datasets. 
For example, in user behavior analysis from clickstreams, each item represents one action in user's click stream ({\eg}, $\mathcal{I}=\{$\texttt{search}, \texttt{select}\}, where $r=2$). Similarly in DNA sequencing, each item represents one canonical base ({\eg}, $\mathcal{I}\!=\!\{\texttt{A},\! \texttt{T},\! \texttt{G},\! \texttt{C}\}$, where $r=4$). 

There are dependencies between items in a sequence. Without loss of generality, we use the one-hot encoding of $S_k$, denoted as $ \mathbf{S}_k = \big(\vec{\alpha}_k^{(1)}, \cdots, \vec{\alpha}_k^{(l_k)}\big) \in \mathbb{R}^{l_k \times r} $
where each item $\vec{\alpha}_k^{(t)}\in \mathbb{R}^r$ in $\mathbf{S}_k$ is a one-hot vector corresponding to the original item $\alpha_k^{(t)}$ in the sequence $S_k$. 

Additionally, each sequence is associated with a set of \textit{attributes}. Each attribute value can be either categorical or numerical. The attribute values are denoted using a vector $\mathbf{x}_k \in \mathbb{R}^{u}$, where $u$ is the number of attributes in $\mathbf{x}_k$. For example, in a dataset where each instance has two attributes ``\texttt{IP}'' and ``\texttt{OS}'', $u=2$. 

With the attributes and sequences, we now formally define the \textit{attributed sequences} (Def.~\ref{attributed-sequence}) and the \textit{attribute-sequence dependencies} (Def.~\ref{def-asr}).

\begin{definition}[Attributed Sequence]{\rm
    \label{attributed-sequence}
    Given a vector of attribute values $\mathbf{x}_k$ and a sequence $\mathbf{S}_k$, an attributed sequence $J_k\! =\! (\mathbf{x}_k,\! \mathbf{S}_k)$ is an ordered pair of the attribute value vector $\mathbf{x}_k$ and the sequence $\mathbf{S}_k$.}
\end{definition}

\begin{definition}[Attribute-Sequence Dependencies]{\rm
\label{def-asr}
Given an attributed sequence $J_k \!=\! (\!\mathbf{x}_k,\! \mathbf{S}_k\!)$, the log likelihood of $J_k$ is $\log \Pr(\!\mathbf{x}_k,\! \mathbf{S}_k\!)$.}%
\end{definition}

\subsection{Problem Definition} %
The goal of attributed sequence embedding is to learn an embedding function that transforms each attributed sequence with a variable-length sequence of categorical items and a set of attributes into a compact representation in the form of a vector. However, these representations are only valuable if an embedding function is capable of learning all three types of dependencies. Hence, given a set of attributed sequences, we define the learning objective of the embedding function as a minimization of the aggregated negative log likelihood of all three types of dependencies. 

\begin{definition}[\small Attributed Sequence Embedding.]{\rm
\label{def-main}
Given a dataset of attributed sequences $\mathcal{J}=\{J_1, \cdots, J_n\}$, the problem of attributed sequence embedding is to find an embedding function $\Theta$ with a set of parameters (denoted as $\theta$) that produces embeddings for $J_k$ in the form of vectors. The problem is formulated as:
\begin{equation}
\label{equation-main}
    {\minimize_{\theta}}\sum_{k=1}^{n} 
    \sum_{t=1}^{l_k}-\log\Pr \left(\vec{\alpha}_k^{(t)} | \vec{\beta}_k^{(t)}, \mathbf{x}_k\right)
\end{equation}
\noindent where $\vec{\beta}_k^{(t)}\! = \!\left(\! \vec{\alpha}_k^{(t-1)},\! \cdots,\! \vec{\alpha}_k^{(1)}\right), \forall t = 2,\! \cdots,\! l_k$ represents the items prior to $\vec{\alpha}_k^{(t)}$ in the sequence. }%
\end{definition}
Our problem can be interpreted as: we want to minimize the prediction error of the $\vec{\alpha}_k^{(t)}$ in each attributed sequence given the attribute values $\mathbf{x}_k$ and all the items prior to $\vec{\alpha}_k^{(t)}$. 
\section{The \sysname Framework}
\label{the-model}

\subsection{Attribute Network} 
\label{rep-attr}
Fully connected neural network~\cite{liou2014autoencoder} is capable of modeling the dependencies of the inputs and at the same time reduce the dimensionality. 
Fully connected neural network has been widely used ~\cite{liou2008modeling, liou2014autoencoder, phan2016differential} for unsupervised data representations learning, including tasks such as dimensionality reduction and generative data modeling. 
With the high-dimensional sparse input attribute values $\mathbf{x}_k \in \mathbb{R}^u$, it is ideal to use such a network to learn the attribute dependencies. We design our attribute network as: 
\begin{equation}
    \label{autoencoder-formula}
    \begin{split}
\vspace{-2pt}
        \mathbf{V}_k^{(1)} &= \rho\left(\mathbf{W}_A^{(1)}\mathbf{x}_k + \mathbf{b}_A^{(1)}\right)\\[-10pt]
        &\vdots\\[-4pt]
        \mathbf{V}_k^{(M+1)} &= \sigma\left(\mathbf{W}_A^{(M+1)}\mathbf{V}_k^{(M)} + \mathbf{b}_A^{(M+1)}\right)\\[-10pt]
        &\vdots\\[-4pt]
        \widehat{\mathbf{x}_{k}} &= \sigma\left(\mathbf{W}_A^{(2M)}\mathbf{V}_k^{(2M-1)} + \mathbf{b}_A^{(2M)}\right)
    \end{split}
\end{equation}
where $\rho$ 
and $\sigma$ 
are two activation functions. 
In this attribute network, we use the \texttt{ReLU} function proposed in~\cite{nair2010rectified} (defined as $\rho(z) = \max(0,z)$) and \texttt{sigmoid} function (defined as $\sigma(z) = \frac{1}{1 + e^{-z}}$). %

The attribute network is an encoder-decoder stack with $2M$ layers, where the first $M$ layers composed of the \textit{encoder} while the next $M$ layers work as the \textit{decoder}. 
With $d_M$ hidden units in the $M$-th layer, the input attribute vector $\mathbf{x}_k \in \mathbb{R}^u$ is first transformed to $\mathbf{V}_k^{(M)}\in \mathbb{R}^{d_M}, d_M \ll u$ by the encoder. Then the decoder attempts to reconstruct the input and produce the reconstruction result $\widehat{\mathbf{x}_k}\in \mathbb{R}^{u}$. 
An ideal attribute network should be able to reconstruct the input from the $\mathbf{V}_k^{(M)}$. 
The smallest attribute network is built with $M=1$, where there are one layer of encoder and one layer of  decoder. 
\subsection{Sequence Network}
\label{rep-multimodal}

The proposed sequence network is a variation of the long short-term memory model (LSTM)~\cite{hochreiter1997long}. The sequence network takes advantage of the conventional LSTM to learn the dependencies between items in sequences. \cite{hochreiter1997long} defines the conventional LSTM model is defined as 
\begin{equation}
\label{seqnet-formula-a}
    \begin{split}
    \mathbf{i}_k^{(t)} &= \sigma\left(\mathbf{W}_i\vec{\alpha}_k^{(t)} + \mathbf{U}_i\mathbf{h}_k^{(t-1)} + \mathbf{b}_i\right) \\[-3pt]
    \mathbf{f}_k^{(t)} &= \sigma\left(\mathbf{W}_f\vec{\alpha}_k^{(t)} + \mathbf{U}_f\mathbf{h}_k^{(t-1)} + \mathbf{b}_f\right) \\[-3pt]
    \mathbf{o}_k^{(t)} &= \sigma\left(\mathbf{W}_o\vec{\alpha}_k^{(t)} + \mathbf{U}_o\mathbf{h}_k^{(t-1)} + \mathbf{b}_o\right) \\[-3pt]
    \mathbf{g}_k^{(t)} &= \sigma\left(\mathbf{W}_g\vec{\alpha}_k^{(t)} + \mathbf{U}_g\mathbf{h}_k^{(t-1)} + \mathbf{b}_g\right) \\[-3pt]
    \mathbf{c}_k^{(t)} &= \mathbf{f}_k^{(t)} \odot \mathbf{c}_k^{(t-1)} + \mathbf{i}_k^{(t)} \odot \mathbf{g}_k^{(t)} \\[-3pt]
    \mathbf{h}_k^{(t)} &= \mathbf{o}_k^{(t)} \odot \tanh\left(\mathbf{c}_k^{(t)}\right) \\[-4pt]
    \end{split}
\end{equation}
where $\odot$ denotes element-wise product, $\sigma$ is a \texttt{sigmoid} activation function, $\mathbf{i}_k^{(t)}, \mathbf{f}_k^{(t)}, $ $\mathbf{o}_k^{(t)}$ and $\mathbf{g}_k^{(t)}$ are the internal gates of an LSTM. The cell states (denoted as $\mathbf{c}_k^{(t)}$) and hidden states (denoted as $\mathbf{h}_k^{(t)}$), which store the information of the sequential data, are two important components in the LSTM model. Without loss of generality, we denote the dimension of the cell states and the hidden states as $d$. 

\textbf{Integration of Attribute Network and Sequence Network. }Different from the conventional LSTM, our proposed sequence network also accepts the output from the attribute network to condition the sequence network. In particular, we have redesigned the function of the \textit{hidden states} to integrate the information from the attribute network by conditioning the sequence network at the first time step as 
\begin{equation}
    \small
\label{seqnet-formula-b}
    \mathbf{h}_k^{(t)} = \mathbf{o}_k^{(t)} \odot \tanh\left(\mathbf{c}_k^{(t)}\right) + \mathds{1}(t=1)\odot\mathbf{V}_k^{(M)} 
\end{equation}
This integration requires the attribute network and the sequence network have the same number of the hidden units (\ie, $d_M = d$).

Since the attributed sequences are unlabeled, we designed the sequence network to predict \textit{the next item in the sequence} as the training strategy. The prediction is carried out by designing an output layer applying a \texttt{softmax} function on the hidden states as
\begin{equation}
    \small
\label{seqnet-formula-c}
    \mathbf{y}_k^{(t)} = \delta\left(\mathbf{W}_y\mathbf{h}_k^{(t)} + \mathbf{b}_y\right)
    \vspace{-2pt}
\end{equation}
where $\mathbf{y}_k^{(t)} \in \mathbb{R}^{r}$ is the predicted next item in sequence computed using \texttt{softmax} function, $\mathbf{W}_y$ and $\mathbf{b}_y$ are the weights and bias of this output layer. With the \texttt{softmax} activation function, the $\mathbf{y}_k^{(t)}$ can be interpreted as the probability distribution over $r$ items. 

\subsection{Training} 
\label{sec-backprop}
\subsubsection{Training Objectives}
We use two different learning objectives for attribute network and sequence network targeting at the unique characteristics of attribute and sequence data. 

\begin{enumerate}
\item Attribute network aims at minimizing the differences between the input and reconstructed attribute values. The learning objective function of attribute network is defined as
\begin{equation}
    \small
    \label{loss-att-net}
    L_{A} = \|\mathbf{x}_k - \widehat{\mathbf{x}_k}\|_2^2
\end{equation}
\item Sequence network aims at minimizing log likelihood of incorrect prediction of the next item at each time step. Thus, the sequence network learning objective function can be formulated using categorical cross-entropy as
\begin{equation}
    \label{loss-seq-net}
    \small
    L_{S} = -\sum_{t=1}^{l_k} \vec{\alpha}_k^{(t)}\log\mathbf{y}_k^{(t)}
\end{equation}
\end{enumerate}
\subsubsection{Embedding}

After the model is trained, we use the parameters in attribute network and sequence network to embed each attributed sequence. 
Specifically, the attributed sequences are used as inputs to the \textit{trained} model only with the one forward pass, where the parameters within the model remain unchanged. 
After the last time step for an attributed sequence $\mathbf{S}_k$, the cell state of sequence network, namely $\mathbf{c}_k^{(l_k)}$, is used as the embedding of $\mathbf{S}_k$. 

\section{Experimental Evaluation}
\label{experiments}
In this section, we evaluate \sysname framework using real-world application logs from \coname and public datasets from Wikispeedia~\cite{west2012human,west2009wikispeedia}. 
We evaluate the quality of embeddings generated by different methods by measuring the performance of outlier detection algorithms using different embeddings. %

\subsection{Experimental Setup}
\subsubsection{Data Collection} 
We use four datasets in our experiments: two from \coname application log files and two from Wikispeedia\footnote{Personal identity information is not collected for privacy reasons. }. The numbers of attributed sequences in all four datasets vary between $\sim$58k and $\sim$106k. 

\begin{itemize}
    \item \textsc{AMS-A/B}: We extract $\sim$164k instances from log files of an Amadeus internal application. Each record is composed of a profile containing information ranging from system configurations to office name, and a sequence of functions invoked by click activities on the web interface. There are 288 distinct functions, 57,270 distinct profiles in this dataset. The average length of the sequences is 18. 
    \item \textsc{Wiki-A/B}: This dataset is sampled from Wikispeedia dataset. Wikispeedia dataset originated from a human-computation game, called Wikispeedia~\cite{west2009wikispeedia}. We use a subset of $\sim$3.5k paths from Wikispeedia with the average length of the path as 6. We also extract 11 sequence context (\eg, the category of the source page, average time spent on each page) as attributes. 
\end{itemize}
\begin{figure*}[t]
    \centering
    \includegraphics[page=1, width=\figwidth\linewidth]{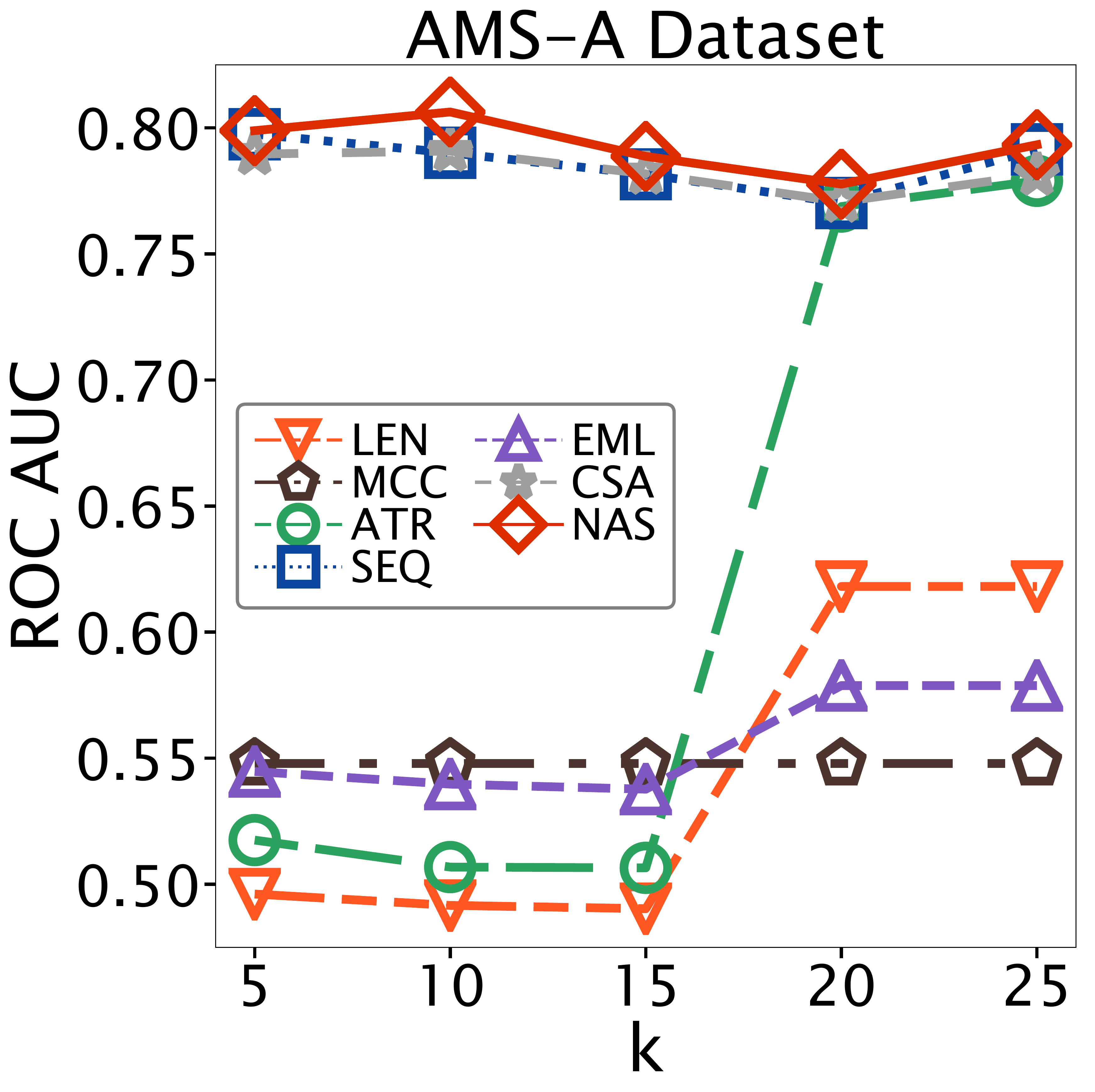}
    \includegraphics[page=2, width=\figwidth\linewidth]{auc_k.pdf}
    \includegraphics[page=3, width=\figwidth\linewidth]{auc_k.pdf}
    \includegraphics[page=4, width=\figwidth\linewidth]{auc_k.pdf}     
    \caption{Parameter sensitivity to different $k$ values. It is shown that the embeddings generated by \sysname always have the best performance under different $k$ values. }
    \label{exp-auc-k}
\end{figure*}
\begin{figure*}[t]
\centering
    \includegraphics[page=1, width=\figwidth\linewidth]{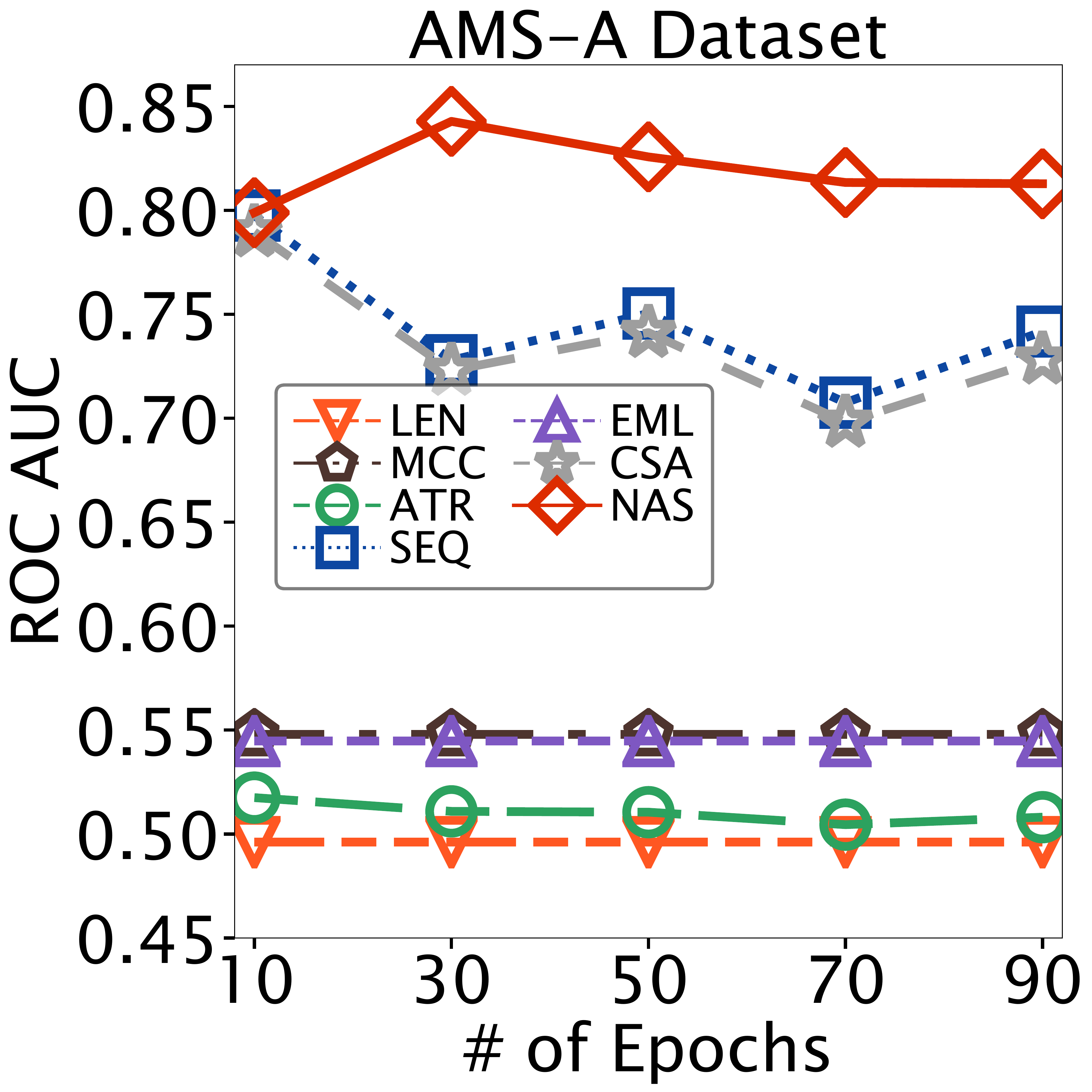}
    \includegraphics[page=2, width=\figwidth\linewidth]{auc_epoch.pdf}
    \includegraphics[page=3, width=\figwidth\linewidth]{auc_epoch.pdf}
    \includegraphics[page=4, width=\figwidth\linewidth]{auc_epoch.pdf}         
    \caption{Performance comparisons using outlier detection tasks. The embeddings generated by \sysname can always achieve the best performance compared to baseline methods when the number of training epochs increases. }
    \label{exp-auc-epoch}
\end{figure*}

\begin{figure}[t]
    \centering
    \includegraphics[width=0.75\linewidth]{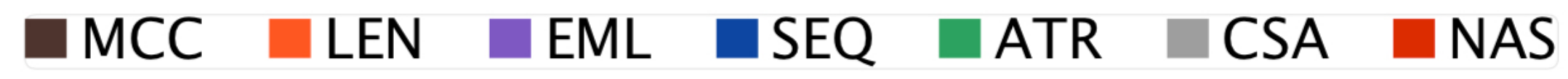}
    
    \begin{subfigure}[t]{0.45\linewidth}
        \centering
        \includegraphics[page=1, width=\linewidth]{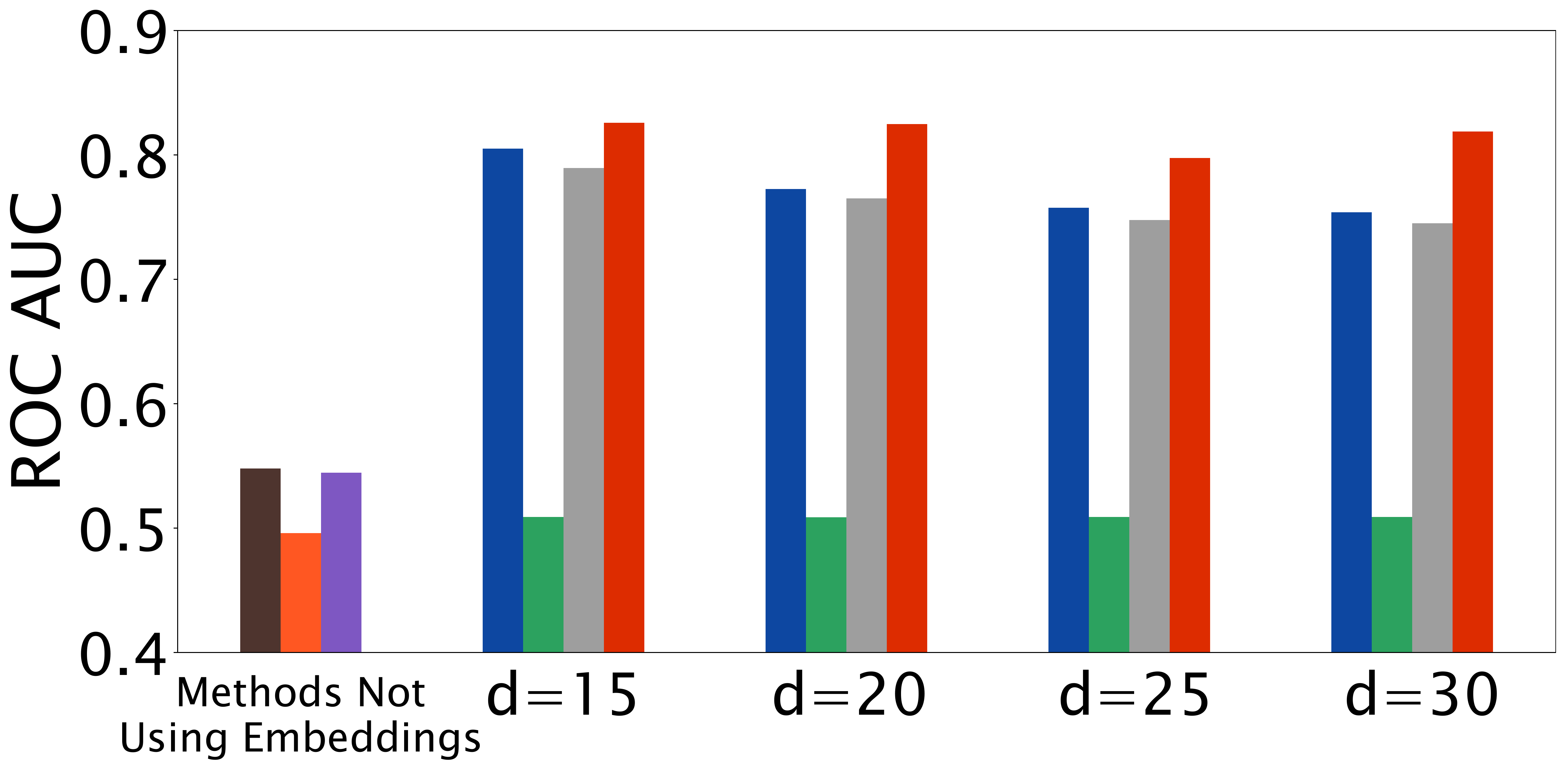}
        \caption{\textsc{AMS-A} Dataset}
        \label{fig-outlier-perf-auc-a}
    \end{subfigure}
    \begin{subfigure}[t]{0.45\linewidth}
        \centering
        \includegraphics[page=2, width=\linewidth]{auc_perf_bar.pdf}
        \caption{\textsc{AMS-B} Dataset}
        \label{fig-outlier-perf-auc-b}
    \end{subfigure}

    \begin{subfigure}[t]{0.45\linewidth}
        \centering
        \includegraphics[page=3, width=\linewidth]{auc_perf_bar.pdf}
        \vspace{-5mm}
        \caption{\textsc{Wiki-A} Dataset}
        \label{fig-outlier-perf-wiki-A}
    \end{subfigure}
    \begin{subfigure}[t]{0.45\linewidth}
        \centering
        \includegraphics[page=4, width=\linewidth]{auc_perf_bar.pdf}
        \vspace{-5mm}
        \caption{\textsc{Wiki-B} Dataset}
        \label{fig-outlier-perf-wiki-b}
    \end{subfigure}
        \vspace{-3mm}
    \caption{The performance of $k$-NN outlier detection ($k=5$). The \textit{methods not using embeddings} are placed on the left. We vary the number of dimensions on the right. The higher score is better. We observe that the combinations of $k$-NN and \sysname embeddings have the best performance on the four datasets. }
    \label{fig-outlier-perf}
    \vspace{-5mm}
\end{figure}

\subsubsection{Compared Methods} 

To study \sysname performance on attributed sequences in real-world applications, we use the following compared methods in our experiments. 
\begin{itemize}
\item %
\textsc{LEN}~\cite{akata2013label}: The attributes are encoded and directly used in the mining algorithm. 
\item %
\textsc{MCC}~\cite{bernhard2016clickstream}: %
\textsc{MCC} uses the sequence component of attributed sequence as input and produces log likelihood for each sequence.
\item %
\textsc{SEQ}~\cite{sutskever2014sequence}: Only the sequence inputs are used by an LSTM to generate fixed-length embeddings. %
\item %
\textsc{ATR}~\cite{wang2014generalized}: A two-layered fully connected neural network is used to generate attribute embeddings.
\item %
\textsc{EML}\cite{yager2014probabilistically}: Aggregate \textsc{MCC} and \textsc{LEN} scores.
\item %
\textsc{CSA}~\cite{ngiam2011multimodal}: The attribute embedding and the sequence embedding are independently generated by \textsc{ATR} and \textsc{SEQ}, then concatenated together. %
\end{itemize} 

\subsubsection{Network Parameters}
Following the previous work in~\cite{glorot2010understanding}, we initialize weight matrices $\mathbf{W}_A$ and $\mathbf{W}_S$ using the uniform distribution. The recurrent matrix $\mathbf{U}_S$ is initialized using the orthogonal matrix as suggested by~\cite{saxe2013exact}. All the bias vectors are initialized with zero vector $\pmb0$. We use stochastic gradient descent as optimizer with the learning rate of 0.01. We use a two-layer encoder-decoder stack as our attribute network. 

\subsection{Outlier Detection Tasks}
\label{outlier}
We use outlier detection tasks to evaluate the quality of embeddings produced by \sysnamenospace. We select $k$-NN outlier detection algorithm as it has only one important parameter (\ie, the $k$ value). We use ROC AUC as the metric in this set of experiments. 

For each of the \textsc{AMS-A} and \textsc{AMS-B} datasets, we ask domain experts to select two users as inlier and outlier. These two users have completely different behaviors ({\ie}, sequences) and metadata ({\ie}, attributes). The percentages of outliers in \textsc{AMS-A} and \textsc{AMS-B} are 1.5\% and 2.5\% of all attributed sequences, respectively. 
For the \textsc{Wiki-A} and \textsc{Wiki-B} datasets, each path is labeled based on the category of the source page. Similarly to the previous two datasets, we select paths with two labels as inliers and outliers where the percentage of outlier paths is 2\%. The feature used to label paths is excluded from the learning and embedding process. 

\textbf{Performance.} 
The goal of this set of experiments is to demonstrate the performance of outlier detection using all our compared methods. Each method is trained with all the instances. %
For \textsc{SEQ}, \textsc{ATR} and \sysnamenospace, the number of learning epochs is set to 10 and we vary the number of embedding dimensions $d$ from 15 to 30. 
We set $k=5$ for outlier detection tasks for \textsc{LEN}, \textsc{SEQ}, \textsc{ATR}, \textsc{CSA} and \sysnamenospace. Choosing the \textit{optimal} $k$ value in the outlier detection tasks is beyond the scope of this work, thus we omit its discussions. We summarize the performance results in Fig.~\ref{fig-outlier-perf}. 

\textbf{Analysis.} We find that the results based on the embeddings generated by \sysname are superior to other methods. 
That is, \sysname maximally outperforms other state-of-the-art algorithms by 32.9\%, 27.5\%, 44.8\% and 48\% on \textsc{AMS-A}, \textsc{AMS-B}, \textsc{Wiki-A} and \textsc{Wiki-B} datasets, respectively. It is worth mentioning that \sysname outperforms a similar baseline method \textsc{CSA} by incorporating the information of attribute-sequence dependencies. 

\textbf{Parameter Study}
There are two \textit{key} parameters in our evaluations, \ie, $k$ value for the $k$-NN algorithm and the learning epochs. 

In Fig. \ref{exp-auc-k}, we first show that the embeddings (dimension $d=15$) generated by our \sysname assist $k$-NN outlier detection algorithm to achieve superior performance under a wide range of $k$ values ($k=5, 10, 15, 20, 25$). We omit the detailed discussions of selecting the optimal $k$ values due to the scope of this work. 

Fig. \ref{exp-auc-epoch} presents the results when we fix $k=5, d=15$ and vary the number of epochs in the learning phase. We notice that compared to its competitor, the embeddings generated by \sysname can achieve a higher AUC even with a relatively fewer number of learning epochs. Compared to other neural network-based methods (\ie, \textsc{SEQ}, \textsc{ATR} and \textsc{CSA}), \sysname have a more stable performance. The \sysname performance gain is not due to the advantage of using both attributes and sequences, but because of taking the various dependencies into account, as the other two competitors (\ie, \textsc{CSA} and \textsc{EML}) also utilize the information from both attributes and sequences.

\section{Related Work}
\label{related}
\textbf{Sequence Mining.} %
Many sequence mining work focuses on frequent sequence pattern mining. 
Recent work in~\cite{miliaraki2013mind} targets finding subsequences of possible non-consecutive actions constrained by a gap within sequences.~\cite{egho2015parameter} aims at solving pattern-based sequence classification problems using a parameter-free algorithm from the model space. It defines rule pattern models and a prior distribution on the model space. 
\cite{fowkes2016subsequence} builds a subsequence interleaving model for mining the most relevant sequential patterns. %

\textbf{Deep Learning.} 
Sequence-to-sequence learning in~\cite{sutskever2014sequence} uses long short-term memory model in machine translation. %
The hidden representations of sentences in the source language are transferred to a decoder to reconstruct in the target language. The idea is that the hidden representation can be used as a compact representation to transfer sequence similarities between two sequences. 
Multi-task learning in~\cite{luong2015multi} examines three multi-task learning settings for sequence-to-sequence models that aim at sharing either an encoder or decoder in an encoder-decoder model setting. Although the above work is capable of learning the dependencies within a sequence, none of them focuses on learning the dependencies between attributes and sequences. This new bipartite data type of attributed sequence has posed new challenges of heterogeneous dependencies to sequence models, such as RNN and LSTM.  Multimodal deep neural networks~\cite{karpathy2015deep, ngiam2011multimodal, xu2015show} is designed for information sharing across multiple neural networks, but none of these work focuses on our attributed sequence embedding problem.

\section{Conclusion}
\label{conclusion}
In this paper, we study the problem of \textit{unsupervised attributed sequences embedding}. 
Different from conventional feature learning approaches, which work on either sequences or attributes without considering the \textit{attribute-sequence dependencies}, we identify the three types of dependencies in attributed sequences. We propose a novel framework, called \sysnamenospace, to learn the heterogeneous dependencies and embed unlabeled attributed sequences. 
Empirical studies on real-world tasks demonstrate that the proposed \sysname effectively boosts the performance of outlier detection tasks compared to baseline methods.

\bibliographystyle{IEEEtran}
\bibliography{ref.bib}
\end{document}